\documentclass[conference]{IEEEtran}
\IEEEoverridecommandlockouts
\usepackage{cite}
\usepackage{amsmath,amssymb,amsfonts}
\usepackage{graphicx}
\usepackage{textcomp}
\usepackage{xcolor}
\def\BibTeX{{\rm B\kern-.05em{\sc i\kern-.025em b}\kern-.08em
    T\kern-.1667em\lower.7ex\hbox{E}\kern-.125emX}}
    
\makeatletter
\newcommand{\linebreakand}{%
  \end{@IEEEauthorhalign}
  \hfill\mbox{}\par
  \mbox{}\hfill\begin{@IEEEauthorhalign}
}
\makeatother

\usepackage{amsmath}
\usepackage{amsthm}
\usepackage{amsfonts}
\usepackage{enumerate}
\usepackage{amsfonts}
\usepackage{algorithm}
\usepackage{algpseudocode}
\usepackage[numbers]{natbib}
\usepackage{graphicx}
\usepackage{xcolor}
\usepackage{enumitem}
\usepackage{multirow}
\usepackage{flushend}

\newtheorem{definition}{Definition}

\usepackage{upquote,textcomp}
\newcommand{\upquote}{\text{\textquotesingle}}

\begin{document}
\title{IPProtect: protecting the intellectual property of visual datasets during data valuation}
\author{\IEEEauthorblockN{Gursimran Singh}
\IEEEauthorblockA{\textit{Huawei Technologies Canada} \\
Burnaby, Canada \\
gursimran.singh1@huawei.com}
\and
\IEEEauthorblockN{Chendi Wang}
\IEEEauthorblockA{\textit{Huawei Technologies Canada} \\
Burnaby, Canada
\\
chendi.wang1@huawei.com}
\and
\IEEEauthorblockN{Ahnaf Tazwar}
\IEEEauthorblockA{\textit{University of British Columbia} \\
Vancouver, Canada\\
tazwarahnaf22@gmail.com}
\linebreakand
\IEEEauthorblockN{Lanjun Wang}
\IEEEauthorblockA{\textit{Tianjin University} \\
Tianjin, China \\
wanglanjun@tju.edu.cn}
\and
\IEEEauthorblockN{Yong Zhang}
\IEEEauthorblockA{\textit{Huawei Technologies Canada} \\
Burnaby, Canada\\
yong.zhang3@huawei.com}
}

\maketitle

\begin{abstract}
Data trading is essential to accelerate the development of data-driven machine learning pipelines. The central problem in data trading is to estimate the utility of a seller's dataset with respect to a given buyer's machine learning task, also known as data valuation. Typically, data valuation requires one or more participants to share their raw dataset with others, leading to potential risks of intellectual property (IP) violations. In this paper, we tackle the novel task of preemptively protecting the IP of datasets that need to be shared during data valuation. First, we identify and formalize two kinds of novel IP risks in visual datasets: data-item (image) IP and statistical (dataset) IP. Then, we propose a novel algorithm to convert the raw dataset into a sanitized version, that provides resistance to IP violations, while at the same time allowing accurate data valuation. The key idea is to limit the transfer of information from the raw dataset to the sanitized dataset, thereby protecting against potential intellectual property violations. Next, we analyze our method for the likely existence of a solution and immunity against reconstruction attacks. Finally, we conduct extensive experiments on three computer vision datasets demonstrating the advantages of our method in comparison to other baselines.
\end{abstract}


\section{Introduction}
Machine learning technology has revolutionized and achieved state-of-the-art performance in many areas like computer vision, natural language processing, and automated driving vehicles. The success of these technologies is imperative on the access to high-quality data, which is often hard and time-consuming to collect \cite{pei2020survey}. Hence, there is an emergent need for a data marketplace, where high-quality datasets can be obtained, in exchange for compensation, with relative ease \cite{liu2021dealer, niu2017trading, an2021crowdsensing}. In a typical data trading scenario, we have data providers (sellers) who offer their curated datasets for a price, data seekers (buyers) who want to buy datasets to boost the model performance on their machine learning task, and optionally a trading platform as a broker to coordinate between two ends of the market participants.

The central problem in a data marketplace is the discovery of the price and the usefulness of various seller datasets for a given buyer task \cite{pei2020survey, liu2021dealer}. A naive way to evaluate a seller dataset's value is to use attributes like the size of the dataset (volume), feature attribute names, class names, etc. However, these attributes do not directly and fully depict the quality of a dataset and hardly correlate closely with the buyer's task performance \cite{pei2020survey}. Hence a more promising solution is \textit{model/utility-based data valuation} where the goal is to estimate the utility a dataset can bring to the buyer's task. In particular, a test dataset provided by the buyer, known as the buyer task dataset $D_B$, is used to measure the utility of a machine learning model trained using a particular seller dataset.
Typically, it requires the buyer and sellers to share their original datasets to the platform for utility estimation \cite{jia2019towards, ghorbani2020distributional}. However, due to the ease of replication of digital assets, a malicious receiver of a dataset can easily store, sell or extract value from the dataset, hence violating the intellectual property (IP) of the dataset owner.

In a broad sense, intellectual property (IP) corresponds to intangible assets of high value. Arguably, a dataset possesses two types of intellectual property, one drawn from the high value of each data sample individually, and the other owing to the statistical value of a well-curated dataset as a whole \cite{margoni2018artificial}. We refer to the former as data-item IP and the latter as statistical IP.
For instance, consider a labeled dataset containing pictures of plants with certain rare diseases. Such a dataset contains two pieces of potentially-tradeable information of high value. The first, data-item IP, arises due to the copyrightable visual information in plant pictures. The second, statistical IP, is due to the high value of the curated dataset which can be used, for example, to train a machine learning-based diagnostic model for plant diseases. Although intellectual property is protected by law, it is often challenging to conclusively demonstrate its violation, especially in the case of digital assets in artificial intelligence \cite{li2020open, hu2018new, zhang2018protecting}. Hence, there is a need to preemptively protect intellectual property before sharing the assets with potential adversaries to preclude them from illegitimately using the high-value assets.

In this work, we propose a novel task of secure data valuation, where the goal is to do accurate data valuation while protecting the intellectual property of all market participants. We limit the focus of the task to the setting where only the buyer is required to share a version of his dataset directly with the sellers for data valuation. Hence, in this setting, the specific goal of our task is to protect the IP of the shared version of the buyer's data while allowing accurate data valuation. On the other hand, our setting trivially provides protection to sellers as they are not required to share their high-value assets (datasets and trained model) before the trade is finalized. Further, we limit the focus of this work to the supervised image classification task and leave extending the proposed approach to other domains for future research. Hence, in this work, the data-item IP originates from the rich visual content in pictures, thereby also referred to as image IP.

Secure data valuation is a challenging problem due to two main reasons. First, the concurrent goals of intellectual property protection and accurate data valuation using the shared version of buyer's data require precise information filtering. To achieve these goals, this filtering is expected to reject all irrelevant information except that is needed for the utility estimation on the seller models. As stated in Section \ref{sec:secure_data_valuation_problem}, these combined goals result in an intractable optimization that cannot be directly solved using existing optimization tools. Second, any attempt at precise information filtering has to be conditioned on the seller models, which are not available at the buyer's end. This is because, in our setting, the sellers do not share their high-value assets with the buyer, for the risk of being stolen, before the trade has been finalized.

In this paper, we make four main contributions as follows.
\begin{itemize}
    \item We define the novel task of secure data valuation, which does not require blind trust between the sellers, the buyers, and the trading platform. We identify two novel security risks involving potential intellectual property violations, namely data-item IP and statistical IP violations, which are serious concerns in the data marketplace.
    \item We propose IPProtect, a novel technique that protects against potential intellectual property violations. Our approach aims to extract only the relevant information required for utility estimation concerning seller models evaluated on the buyer task dataset. By rejecting irrelevant information, our approach effectively limits information transfer, hence protecting against potential intellectual property violations. 
    \item We propose a novel optimization based on the concept of statistical generalization in machine learning. It allows accurate utility preservation on seller models, which have not been observed, given only sampling access to the distribution from which these models are assumed to be sampled. We provide an analysis of the existence of a solution and empirical validation of the statistical generalization of our novel optimization.
    \item We demonstrate the superior performance of our approach on utility preservation and security preservation when compared to baseline solutions on three benchmark computer vision datasets. We establish immunity to reconstruction-type attacks by providing an analysis of the mathematical hardness of the attack along with empirical experiments.
\end{itemize}
\section {Related Work}
Detection or prevention of infringement of intellectual property (IP) of high-value assets in machine learning is a problem of great interest \cite{drexl2019technical, bertino2005privacy, ganesh2019watermark, zhang2018protecting, devarapalli2018machine, chakraborty2020hardware}. A majority of existing works \cite{li2020open, hu2018new, zhang2018protecting} aim to embed watermarks into machine learning models, which can subsequently be detected for potential IP infringements. However, the watermarking-based techniques are aimed toward the IP protection of models \cite{boenisch2020survey}, while our approach targets the IP protection of datasets. Additionally, watermarking only serves to provide detection of potential IP violations, which is limited as the subsequent legal recourse, is often cumbersome and time-consuming. On the other hand, our approach aims to preemptively prevent adversaries to steal intellectual property in the first place by limiting arbitrary misuse of datasets.

Secure data valuation, which requires IP protection of datasets, is a novel task that has not been systematically studied in the literature. Hence, we review existing work mainly in the domain of private data synthesis \cite{chen2020gs,harder2021dp,cao2021don,huang2020instahide,liu2020datamix} and encryption-based methods \cite{gilad2016cryptonets, koutsos2021agora, yonetani2017privacy}, both of which aim to hide information in datasets. In general, private data synthesis approaches aim to synthesize a sanitized dataset, for secure sharing, with a focus on hiding all identification information to protect privacy \cite{dwork2014algorithmic}. On the other hand, the focus of IP protection is to limit the sanitized dataset's capabilities to be illegitimately traded in the market. Specifically, it aims to synthesize images devoid of the visual appeal of original images and prevent datasets to be used for arbitrary statistical analysis, thereby, preventing potential IP infringements.

DP-GAN-based approaches \cite{chen2020gs,harder2021dp,cao2021don} use DP-based techniques to train a generative adversarial network (GAN) with the original dataset, followed by conditional sampling to generate a sanitized dataset. The key idea is that, by post-processing theorem \cite{dwork2014algorithmic}, a dataset sampled from a GAN trained with DP, is also expected to be DP. Although DP offers to hide the membership information of a particular data item, it is not designed to protect attribute-level visual information or mask dataset-level statistical information. As a result, they are not effective in protecting either image or statistical intellectual property.

Similarly, instance-hiding approaches \cite{huang2020instahide, liu2020datamix}, synthesize private images by weighted averaging of a random set of images with random weights. In some cases, this is followed by applying a random pixel-wise mask to further boost security \cite{huang2020instahide}. The resulting mixed-up images provide "lightweight" privacy by obfuscating images and making them visually incomprehensible. The visual obfuscation can potentially protect against image IP violations, however, recently, several reconstruction-type attacks \cite{carlini2021private, luo2021fusion} have been proposed which can recover high-fidelity originals even for state-of-the-art variants like InstaHide \cite{huang2020instahide}. On top of that, these methods are not designed to protect statistical information, rendering them ineffective for statistical IP protection.

On the other hand, certain cryptographic setups \cite{gilad2016cryptonets, koutsos2021agora, yonetani2017privacy} can be employed for preventing IP violations by restricting information access through encryption.
For instance, homomorphic encryption \cite{gilad2016cryptonets, yonetani2017privacy} can be used to perform secure inference on seller models while preventing sellers from decrypting and misusing the buyer's data. In addition, functional encryption \cite{koutsos2021agora} can be used to limit the allowed computations of the buyer data on the seller side, preventing arbitrary IP violation risks. In theory, these methods can provide effective protection against image and statistical intellectual property violations, however, various practical challenges drastically limit their applicability to modern computation-demanding machine/deep learning pipelines \cite{boulemtafes2020review, huang2020instahide}. 
Another recent work \cite{xue2021protect} uses symmetric encryption to restrict unauthorized access to a publicly released dataset with a more explicit aim to protect IP. However, it lacks a clear definition of IP and, is not extendable to the setup of secure data valuation task.

In contrast to cryptographic methods, our framework is not designed to hide all information in data samples, instead, it selectively filters information such that irrelevant information on the IP aspect is forgotten while, at the same time, preserving necessary information for accurate data valuation.
\section{Problem formulation} \label{sec:prob_deb}
In this section, we define intellectual property exposure for visual datasets and introduce the task of secure data valuation. However, before moving to the secure version, we introduce the vanilla (insecure) version of the data valuation task. In a typical data trading scenario that relies on utility-based data valuation, we have a buyer with a target task, represented with a task dataset $D_B := \{(x_k, y_k) | k \in \{1,\ldots, |D_B|\}\}$, which is assumed to be sampled from a distribution $P_B(X,Y)$.
 
On the other hand, we have multiple sellers (say $M$) offering their datasets as candidates to improve performance on the buyer's target task. The $i^{th}$ seller dataset is denoted by $D_{S_i}:= \{(x_k,y_k), k \in \{1,\ldots, |D_{S_i}|\}\}$, where $i\in \{1,\ldots, M\}$. Additionally, we assume both the buyer and seller datasets are designed to solve the same supervised image classification task. For both buyer and seller datasets, $x_k \in \mathbb{R}^{L \times H \times W}$ is the input image and $y_k \in C$ is the class label, where $L$, $H$, $W$, and $C$ represents channels, height, width, and set of class labels, respectively. Finally, there is a trading platform that provides protocols for machine learning pipelines consisting of a model architecture $m_p$ and the algorithm \textit{learn} for training machine learning models.

\begin{definition}[Insecure data valuation task] \label{def:insecure_data_trading}
Given $M$ seller datasets $\{D_{S_i}\}_{i=1}^M$, platform-specified model architecture $m_p$ along with the training algorithm `$learn$', the insecure data valuation task is defined as estimating the utility of each seller dataset evaluated on the buyer task dataset $D_B$ as follows. 
{\small
\begin{align} 
    U_{S_i}(D_B) := \frac{1}{|D_B|}\sum_{k=1}^{|D_B|} \mathbb{I}(\text{argmax } g_{\theta^i} (x_k) == y_k) \label{insecure_util}\\
    \text {where } g_{\theta^i} := \text{learn}(D_{S_i}, m_p) \label{insecure_learn}
\end{align}}
\end{definition}
where $\mathbb{I}$ is an indicator function, and $g_{\theta^i} \in \mathbb{R}^{|C|}$ is the $i^{th}$ seller model. Without loss of generality, in Eq. \eqref{insecure_util}, we have used classification accuracy as the utility function in this work, which can be adapted to other preferable performance metrics. 
Intuitively, seller datasets that are closer to the buyer task distribution $P_B(X,Y)$ are assigned a higher value of utilities. Having solved the above task, the set of utilities $\{U_{S_i}(D_B)\}_{i=1}^M$ can be used for a utility-based data valuation and a fair price discovery for the seller datasets before an agreement of trading deal is reached. Notice that computing Eq \eqref{insecure_util} and Eq. \eqref{insecure_learn} requires simultaneous access to both seller and buyer datasets. However, before the trade is finalized, neither the buyer nor the sellers want to share their high-value assets with each other or the platform due to the potential risk of intellectual property violations defined in Section \ref{sec:ipdefns}, from threats defined in Section \ref{sec:threats}. 

\subsection{IP exposure definitions} \label{sec:ipdefns}
As mentioned earlier, an image dataset contains two pieces of potentially-tradeable information of high value. The first is due to the unique visual content inside individual images and the second is due to the rich statistical information inside the dataset as a whole.

In this section, we propose novel definitions of IP exposure to quantify the risk of potential IP violations against a dataset. We assume an original dataset $D_O := \{(x_k, y_k), k \in \{1,\ldots, |D_O|\}\}$, whose IP is to be protected, and a sanitized version $\Tilde{D}_O := \{(\tilde{x}_k, y_k),$ $ k \in \{1,\ldots, |\Tilde{D}_O|\}\}$, proposed to mask the IP of the original dataset. Additionally, we denote the distribution from which dataset $D_O$ and $\tilde{D}_O$ are sampled as $P_O(X,Y)$ and $\tilde{P}_O(X,Y)$, respectively. Since only the sanitized dataset is to be shared with potential adversaries, the risk of IP violation of the original dataset $D_O$ is dependent on the amount of IP exposed by the sanitized dataset $\tilde{D}_O$. With this aim, we define metrics for evaluating image and statistical IP exposure below.

\subsubsection{Image IP exposure}
Image IP exposure is measured by the perceptual similarity of an image and its sanitized version. The key idea is that a sanitized image, that looks similar to an original image, can be traded for its rich visual content, hence violating the IP of the original image. We use a well-known image quality assessment metric, perceptual loss \cite{johnson2016perceptual}, which deems to correlate well with human visual perception \cite{zhang2018unreasonable, tariq2019analysis}, to define the data-item level image IP.

\begin{definition}[Image IP exposure] \label{image_ip}
Given an oracle to estimate the perceptual distance $\mathbb{D}$ between two images, the image IP exposure is defined as the inverse of the perceptual distance between a pair of original image $x_o$ and its sanitized version $\tilde{x}_o$.

{\small
\begin{align} 
\mathcal{E}_{I}(\tilde{x}_o, x_o) = \frac{1}{\mathbb{D}(\tilde{x}_o, x_o)} \label{def:image_ip}
\end{align}}

\end{definition}
where $\mathbb{D}$ is taken as the perceptual loss \cite {johnson2016perceptual}, which uses representations learned by pretrained deep neural networks to estimate the perceptual distance between a pair of images. Specifically, let $\phi$ be the ImageNet-pretrained VGGNet, perceptual loss is defined as $\mathbb{D}(\tilde{x}_o, x_o) = \frac{\|\phi_j(\tilde{x}_o) - \phi_j(x_o)\|_2^2}{L_j \times H_j \times W_j}$, where $\phi_j$ is the $j^{th}$ convolutional feature map of size $L_j \times H_j \times W_j$.
Further, we define the image IP of a dataset $\mathcal{E}_{ID}(\tilde{D}_O, D_O)$ as the average of the image IP exposure computed for the corresponding pair of images in the original dataset $D_O$ and the sanitized dataset $\tilde{D}_O$.

\subsubsection{Statistical IP exposure}
Statistical IP exposure is based on the usefulness of the sanitized dataset $\tilde{D}_O$ for arbitrary statistical analysis, like model training or model selection. The key idea is that, to protect statistical IP, the receiver of the dataset should not be able to use it to train effective machine learning models or be able to use it for model selection through inference. In order to define it, we measure the average performance of training and inference on a variety of machine learning models (represented by the set $\mathcal{G}$) as below.

\begin{definition}[Training statistical IP exposure]
Given a sanitized dataset $\tilde{D}_O$ and an arbitrary dataset $D_X$, sampled i.i.d from the original data distribution $P_O$, the training statistical IP exposure is defined as the ability of using $\tilde{D}_O$ to train arbitrary machine learning models for effective inference on $D_X$.
{\small
\begin{align} 
    \mathcal{E}_{Tr}(\tilde{D}_O, P_O) = \frac{1}{|\mathcal{G}|}&\sum_m\frac{1}{|D_X|}\sum_k^{|D_X|} \mathbb{I} (\text{argmax } h_m(x_k) == y_k) \label{def:train_stat_ip}\\
    &h_m = \text{learn}(\tilde{D}_O, m) \hspace{3mm} \text{where} \hspace{2mm} m\in \mathcal{G}
\end{align}}
\end{definition}

\begin{definition}[Inference statistical IP exposure]
Given a sanitized dataset $\tilde{D}_O$ and an arbitrary dataset $D_X$, sampled i.i.d from the original data distribution $P_O$, the inference statistical IP exposure is defined as the effectiveness of the sanitized dataset $\tilde{D}_O$ to be used for effective inference on arbitrary machine learning models trained with $D_X$.

{\small
\begin{align} 
    \mathcal{E}_{In}(\tilde{D}_O, P_O) = \frac{1}{|\mathcal{G}|}&\sum_m\frac{1}{|\tilde{D}_O|}\sum_k^{|\tilde{D}_O|} \mathbb{I} (\text{argmax } h_m(x_k) == y_k) \label{def:infer_stat_ip}\\
    &h_m = \text{learn}(D_X, m) \hspace{3mm} \text{where} \hspace{2mm} m\in \mathcal{G}
\end{align}}
\end{definition}

To eliminate bias, we choose a wide variety of models in the set $\mathcal{G}$. This list includes Logistic Regression, Bernoulli Naive Bayes, Gaussian Naive Bayes, Random Forest, Linear SVC, Decision Tree, Linear Discriminant Analysis, ADA Boost, Multi-Layer Perceptron (MLP), Convolutional Neural Network (CNN), Bagging, Gradient Boosting Classifier, and XG Boost.

\subsection{Secure data valuation task} \label{sec:secure_data_valuation_problem}
The insecure data valuation task (Def. \ref{def:insecure_data_trading}) raises risks of potential intellectual property violations as the participants are required to share their original datasets. At a high level, our work aims to design an approximation of vanilla data valuation task Eq \eqref{insecure_util} and \eqref{insecure_learn} without the participants having to share their original datasets. In particular, in our proposed framework, only the buyer is required to share a version of its dataset with the sellers, where the utility estimation can be performed locally. The computed utility information is then relayed to both the platform and the buyer by the sellers, who we assume are less likely to misreport due to the risk of being banned from the platform.

Hence, in our framework, the specific goal of the secure data valuation task is to synthesize the shared version (also referred to as sanitized version) of the buyer dataset $\tilde{D}_B$ with the dual goal of high \textit{utility preservation} and high \textit{security preservation}. High utility preservation requires the seller dataset utilities Eq. \eqref{insecure_util} computed with the shared dataset $\tilde{D}_B$ to be close to those computed with the original dataset $D_B$. Mathematically, for an arbitrary small constant $\delta$, we need $|U_{S_i}(\tilde{D}_B) - U_{S_i}(D_B)| \leq \delta$ for all sellers $i$. High security preservation requires the IP exposed by the shared dataset $\tilde{D}_B$ should be as small as possible in comparison to the baseline exposure of the original dataset $D_B$. Specifically, we need the averaged image IP exposure $\mathcal{E}_{ID}(\tilde{D}_B, D_B)$, the training IP exposure $\mathcal{E}_{Tr}(\tilde{D}_B, P_B)$, and the inference IP exposure $\mathcal{E}_{In}(\tilde{D}_B, P_B)$ to be small.

However, achieving both objectives together is challenging as there is an inherent tradeoff between utility preservation and security preservation. For instance, a shared dataset sampled from the original buyer task distribution $\tilde{D}_B \sim P_B(X,Y)$ provides high utility preservation at the cost of poor security preservation. On the other hand, a shared dataset sampled from a random normal distribution $\tilde{D}_B \sim \mathcal{N}^{L \times H \times W}(0,1)$ provides high security preservation at the cost of very poor utility preservation. The goal of this work is to optimize both the utility and security fronts simultaneously and improve the inherent utility-security preservation tradeoff. An attempt to directly optimize these goals in a multi-objective optimization results in an intractable optimization. In the next section, we propose an alternate proxy optimization to effectively and efficiently optimize these objectives.

\subsection{Threat model} \label{sec:threats}
Our framework does not require the buyer or the sellers to trust each other or the platform. In particular, we require the buyer to only send a shared version of its task dataset $\tilde{D}_B$ to the sellers for utility computation. Hence, potential adversaries include sellers who may illegitimately use the shared dataset $\tilde{D}_B$ for stealing intellectual property. For instance, IP-worthy images in $\tilde{D}_B$ can be sold elsewhere, or the dataset $\tilde{D}_B$ can be used for training a machine learning model or performing model selection, through inference. Additionally, specific to our solution, we assume adversaries have access to the exact data sanitation algorithms used by the buyer. Further, along with the sanitized dataset $\tilde{D}_B$, they have access to an auxiliary dataset $D_A$ which has a distribution that is close to the buyer task data distribution $P_B(X,Y)$. Using this information, an adversary can try reconstruction attacks to recover $D_B$ from $\tilde{D}_B$. Finally, we assume the set of labels $\{y_k\}$, alone, do not possess any intellectual property, and, hence, in the shared dataset $\tilde{D}_B:=\{(\Tilde{x}_k, y_k)\}_{i=1}^{|\tilde{D}_B|}$, the labels are not sanitized.
\section{Method} \label{sec:method}
In this section, we explain our approach, IPProtect, to synthesize the shared (a.k.a sanitized) version of the buyer dataset $\tilde{D}_B$, required by the secure data valuation task. As mentioned earlier, in Section \ref{sec:secure_data_valuation_problem}, directly optimizing the dual goal of high utility preservation and high security preservation is challenging and results in intractable optimization. In our approach, we propose an alternate transformation to convert an original data point $(x_k, y_k) \in D_B$ into a new sanitized data point $(\tilde{x}_k, y_k) \in \tilde{D}_B$, for each data point individually. The transformation $e:(x_k, y_k) \mapsto (\Tilde{x}_k, y_k)$ aims to optimize two proxy goals which indirectly lead to high utility preservation and high security preservation. For the proxy utility preservation goal, we aim that, under all seller models, the projection of the synthesized data point $g_{\theta^i}(\Tilde{x}_k)$, is approximately equal to the projection of the original data point $g_{\theta^i}(x_k)$; here the projection represents a mapping from a data point in the input space $\mathbb{R}^{L \times H \times W}$ to the output-class distribution (logits) $\mathbb{R}^{|C|}$ of a given seller model $g_{\theta^i}$. For the proxy security preservation goal, we aim the sanitized data point $\tilde{x_k}$ to lie close to random normal distribution $\mathcal{N}^{L \times H \times W}(0,1)$. Such a data point ends up being perceptually dissimilar to the original data point, directly resulting in low IP exposure (Def. \ref{def:image_ip}). Consequently, the entire sanitized dataset $\tilde{D}_B$, obtained using the transformation $e$ separately for each data point in $D_B$, also results in low statistical IP as a by-product of this optimization. Using these ideas, for a given $(x_k, y_k) \in D_B$, we formulate the transformation $e$ as the following optimization.  

\begin{align}
\resizebox{.45\textwidth}{!}{
$\underset{\tilde{x}_k}{\operatorname{argmin}} \underbrace{\frac{1}{M}\sum_{i=1}^{M} [ \mathcal{L} ( g_{\theta^i} (\tilde{x}_k), g_{\theta^i} (x_k))]}_{\text{deterministic utility loss}} +  \underbrace{\lambda\text{ max} \{\mathcal{D} (\tilde{x}_k, x_N) -\tau, 0\}}_{\text{security loss (regularisation)}} \label{eq:proxy_sellermodels}$}
\end{align}

where $g_{\theta^i}: X \to \mathbb{R}^C$ is the $i^{th}$ model, among $M$ seller models, $\mathcal{L}$ denotes a loss function for comparing two output class distributions and $\mathcal{D}$ denotes a distance function between the two data points. $\mathcal{L}$ and $\mathcal{D}$ can be any differentiable loss functions or distance functions, which, in our case, we set to $L_2$ distance for both cases. The regularisation anchor $x_N$ denotes a random point sampled from the normal distribution $\mathcal{N}^{L \times H \times W}(0,1)$. $\tau$ represents the radius of the regularisation ball and $\lambda$ controls the weight of the regularisation term in the overall objective function.

The first term, \textit{deterministic utility loss}, in Eq. \eqref{eq:proxy_sellermodels} aims the solution $\tilde{x}_k\upquote$ to have a similar output-class distribution as the original point $x_k$ for all seller models. Mathematically, it ensures $\mathcal{L}(g_{\theta^i}(\tilde{x}_k\upquote), g_{\theta^i}(x_k)) \leq \delta$ $\forall i$, where $\delta$ is a small constant. Practically, the two approximately similar output-class distributions usually end up having the same final prediction verdict, i.e. $\operatorname{argmax} g_{\theta^i}(\tilde{x}_k\upquote) = \operatorname{argmax} g_{\theta^i}(x_k)$ $\forall i$. However, in theory, this is only true for a sufficiently small $\delta$, coupled with sufficiently low entropy of $g_{\theta^i}(x_k) \forall i$. Consequently, by solving the optimization separately for each $x_k \in D_B$, we can intuitively expect the aggregated utility of $\tilde{D}_B$ and $D_B$ to be approximately similar for all seller models $g_{\theta^i}$, resulting in high utility preservation. On the other hand, the \textit{security loss (regularization)} term wants the solution $\tilde{x}_k\upquote$ to be in a ball of radius $\tau$ around a random sample $x_N \sim \mathcal{N}^{L \times H \times W}(0,1)$. This criterion aims that the synthesized $\tilde{x}_k\upquote$ is perceptually different from the original $x_k$, resulting in low image IP exposure. Further, the dataset $\tilde{D}_B$, obtained by synthesizing multiple $\tilde{x}_k\upquote$ close to $\mathcal{N}^{L \times H \times W}(0,1)$, for each $x_k \in D_B$, also ends up having low statistical IP exposure as a by-product. This is because the distribution of the sanitized dataset, consisting of images close to $\mathcal{N}^{L \times H \times W}(0,1)$ is far from the distribution of the original dataset $D_B$, which is a sufficient requirement for low statistical IP exposure. Hence, resulting in high security preservation.

However, directly solving Eq. \eqref{eq:proxy_sellermodels} requires explicit access to seller models which are not available at the buyer's end. So, how can we solve Eq. \eqref{eq:proxy_sellermodels} without knowing $g_{\theta^i}$s in advance? We take inspiration from the concept of generalization in statistical parameter estimation. At a high level, generalization allows a parameterized random variable (e.g. loss), minimized on a finite independent and identically distributed (i.i.d) sample of a distribution, to be upper bounded over the entire distribution with high probability \cite{bousquet2003introduction}. Hence, generalization paves the way for ensuring predictable behavior (e.g. low loss) on samples not known or observed in advance, provided they are sampled from the same distribution. Using this idea, we propose a novel statistical estimation of a data point over a distribution of models. Let's assume a distribution over the parameters of all possible seller models with a known architecture configuration in the data marketplace, and denote it by $\Omega_\theta$. Then, for a given point $(x_k, y_k) \in D_B$, we aim to synthesize $(\tilde{x}_k, y_k) \in \tilde{D}_B$ by estimating the parameters $\tilde{x}_k$ that minimizes the random variable $\mathcal{L} ( f_{\theta} (x_k), f_{\theta} (\tilde{x}_k))$, where $f_\theta \sim \Omega_\theta$. Technically it requires minimizing the expectation, which in practice can be approximated by a finite set $\Omega_{Tr}$ from the distribution $\Omega_\theta$, an idea akin to empirical risk minimization in supervised machine learning \cite{bousquet2003introduction}.

\begin{align}
\resizebox{.48\textwidth}{!}{
$\underset{\tilde{x}_k}{\operatorname{argmin}} \underbrace{\frac{1}{|\Omega_{Tr}|}\sum_{n=1}^{|\Omega_{Tr}|} [ \mathcal{L} ( f_{\theta_n} (\tilde{x}_k), f_{\theta_n} (x_k))]}_{\text{statistical utility loss}} +  \underbrace{\lambda\text{ max} \{\mathcal{D} (\tilde{x}_k, x_N) -\tau, 0\}}_{\text{security loss (regularisation)}} \label{proxy-opt-sum}$}
\end{align}

where the shorthand $f_{\theta_n}$ corresponds to $\Omega_{Tr}[n]$. We hypothesize that the solution $\tilde{x}_k^o$ of Eq. \eqref{proxy-opt-sum} generalizes over all members of the distribution $\Omega_\theta$, under the right conditions for generalization like the sample $\Omega_{Tr}$'s ability to sufficiently express the complexity of the distribution. This implies that, similar to traditional supervised learning \cite{bousquet2003introduction}, for some arbitrary small constant $\delta$, the random variable can be upper bounded $\mathcal{L} ( f_{\theta} (\tilde{x}_k^o), f_{\theta} (x_k))$ $\le \delta$ $\forall f_\theta \sim \Omega_\theta$ with high probability. As a result, we can expect the solution $\tilde{x}_k^o$ to preserve projections for the seller models as well, $\mathcal{L}(g_{\theta^i}(\tilde{x}_k^o), g_{\theta^i}(x_k)) \leq \delta$, with high probability, since $\forall i$ $g_{\theta^i} \sim \Omega_\theta$. The set $\Omega_{Tr}$ is like the training dataset, hence a higher $|\Omega_{Tr}|$ leads to better generalization and, thereby, better utility preservation. However, a higher $|\Omega_{Tr}|$ makes the regularisation term (and security preservation) harder to enforce as the freedom in solution space is restricted due to the increased number of equations in comparison to the parameters. Hence, similar to $\lambda$ and $\tau$, $|\Omega_{Tr}|$ acts as a parameter to control the weight of utility and security terms in Eq. \eqref{proxy-opt-sum}, and consequently, the utility-security preservation tradeoff of our approach. Further, in Section \ref{sec:sol_exists} we provide a detailed analysis of the likely existence of a solution satisfying the utility loss and security loss conditions in Eq. \eqref{proxy-opt-sum}.

In contrast to statistical parameter estimation in supervised learning \cite{bousquet2003introduction}, where the goal is to estimate the parameters of the model $f_\theta$, over a distribution of data points $P(X,Y)$, in our case, the goal is to estimate the sanitized data point $\tilde{x}_k$, over a distribution of models $\Omega_\theta$. Despite the difference, the underlying fundamentals of generalization \cite{bousquet2003introduction} hold in both cases. The estimated parameters, which are $\theta^o$ in the former and $\tilde{x}_k^o$ in the latter case, are \textit{hypothesized to generalize} to unseen samples (validation set), from the corresponding distribution, which is $P(X,Y)$ in the former and $\Omega_\theta$ in the latter case. 
Further, similar to traditional supervised learning \cite{bousquet2003introduction}, we can estimate the generalization performance by sampling another finite set $\Omega_{V}$, the unseen validation set, and computing the empirical generalization $\hat{R}(\Tilde{x}^o) = \frac{1}{|\Omega_{V}|}\sum_{n=1}^{|\Omega_{V}|} [ \mathcal{L} ( f_{\theta_n} (x), f_{\theta_n} (\Tilde{x}^o))]$. This estimation of generalization gives a fair idea of the practical performance of the synthesized dataset $\tilde{D}_B$ in terms of utility preservation even for unseen models sampled from the distribution $\Omega_\theta$. The accuracy of this estimation can be increased by using a larger size of $\Omega_V$, which is guaranteed mathematically by Hoeffding's inequality \cite{zhang2020concentration, bousquet2003introduction}. In Section \ref{sec:stat_generalization}, we conduct a proof-of-the-concept experiment to empirically validate our hypothesis that the solutions of Eq. \ref{proxy-opt-sum} tend to generalize even on unseen (not present in $\Omega_{Tr}$) members of distribution $\Omega_\theta$.

To obtain sets $\Omega_{Tr}$ and $\Omega_V$, we utilize the buyer task dataset to train multiple models to emulate a sample from the distribution $\Omega_\theta$. To ensure a diverse sample, we bootstrap the buyer dataset $D_B$ to train multiple models with random initialization. Further, we use the same platform-specified learning algorithm $learn$ and model architecture $m_p$ as used by all the sellers. Intuitively, we can equate our sampling process, which we call $sample\_models$, to generate models approximately similar to those of seller models $g_{\theta^i}s$. Overall, using the buyer dataset to sample from $\Omega_\theta$ makes sense since the data distributions of the most relevant sellers lie close to the buyer task distribution $P_B(X,Y)$. Although the samples obtained using this process are not necessarily i.i.d samples from $\Omega_\theta$, however, as in many practical supervised learning pipelines, this is not a strict requirement. For instance, a cat vs dog classification model trained with images of certain breeds, can, under modest distribution shifts, transfer to similar looking out-of-distribution breeds \cite{shen2021towards, bickel2009discriminative}. 
In Section \ref{sec:exp_stat_ip}, we empirically observe out-of-distribution generalization in our setup as we obtain good results even when the unseen seller models and the seen seller models (obtained using $sample\_models$) are trained differently.

Using these sampled sets $\Omega_{Tr}$ and $\Omega_V$, we use Eq. \eqref{proxy-opt-sum} to convert the original task dataset $D_B$ into the sanitized version $\tilde{D}_B$. The entire process is described in the Algorithm \ref{alg:proxy_gen}. 

\begin{algorithm}[ht]
  \caption{\small IPProtect data synthesis (Outline)}\label{alg:proxy_gen}
  {\small
  \begin{algorithmic}[1]
    \Require Inputs: buyer task dataset $D_B=\{(x_k,y_k)\}_{k=1}^{|D_B|}$. Parameters: learning rate $\eta_t$, model batch size $B$, sampled train-model set size  $|\Omega_{Tr}|$, sampled val-model set size $|\Omega_V|$, loss-stopping criteria $\zeta$. $T$ is the upper limit of iterations. 
        \State {Initialize $\Tilde{D_B} \gets \phi$, $L_v \gets \phi$} \Comment{$L_v$ is set of validation losses}
        \State {$\Omega_{Tr} \gets \text{sample\_models}(D_B, |\Omega_{Tr}|)$} \Comment{train train-model set}
        \State {$\Omega_{V} \gets \text{sample\_models}(D_B, |\Omega_{V}|)$}   \Comment{train val-model set}
        \For {$(x_k, y_k) \in D_B$}
            \State {$C_k \gets \{f_{\theta}(x_k) | f_{\theta} \in \Omega_{Tr}\}$} \Comment{cache $f_{\theta}(x_k)$ for efficiency}
            \State {Initialize $\Tilde{x}^0 \gets \mathcal{N}^{|\mathcal{X}|}_{(0,1)}$}
    		\For{$t \in [T]$}
        		\State {Randomly sample $\Omega_{Tr}^t \subseteq \Omega_{Tr}$ with probability $B/|\Omega_{Tr}|$}
        		\State {Compute gradient for \textbf{utility} preservation loss}
				\State {$g^t_k =$ {\small $\frac{1}{|\Omega_{Tr}^t|} \sum_{n=1}^{|\Omega_{Tr}^t|} \left[ \frac{\partial \mathcal{L}(f_{\theta_n}(x_k), f_{\theta_n}(\Tilde{x}_k^t))}{\partial f_{\theta_n}(\Tilde{x}_k^t)} . \frac{\partial f_{\theta_n}(\Tilde{x}_k^t)}{\partial \Tilde{x}_k^t} \right]$}} 
        		\State {Compute gradient for \textbf{security} preservation loss}
        		\State {$g^t_k = g^t_k + \frac{\partial \mathcal{D}(x_N, \Tilde{x}_k^t)}{\partial  \Tilde{x}_k^t}$}
        		\State {$\Tilde{x}_k^t = \Tilde{x}_k^{ t-1} - \eta_t g^t_k$} \Comment{update $\Tilde{x}$ from $\Tilde{x}_k^{ t-1}$ to $\Tilde{x}_k^t$}
        		\State {$L_k^t \gets \frac{1}{|\Omega_{V}|}\sum_{n=1}^{|\Omega_{V}|} [ \mathcal{L} ( f_{\theta_n} (x_k), f_{\theta_n} (\Tilde{x}_k^t))]$}  
        		\State {\textbf{if} $|L_k^t - L_k^{t-1}| < \zeta$} \textbf{then} \textit{break} \Comment{stopping criteria}
    		\EndFor
    		\State {Update $\Tilde{D_B} \gets \Tilde{D_B} \cup ({\Tilde{x}_k^t,y_k})$ and $L_v \gets L_v \cup L_k^t$}
        \EndFor
      \State \textbf{Output:} sanitized dataset $\Tilde{D_B}$ and validation losses $L_v$
  \end{algorithmic}}
\end{algorithm}
\section{Analysis}
\subsection{Existence of a solution} \label{sec:sol_exists}
In this section, we provide reasoning for the likely existence of a solution $\tilde{x}_k^o$ simultaneously satisfying the utility loss and security loss constraints in  Eq. \eqref{proxy-opt-sum}. Specifically, we need the solution $\tilde{x}_k^o$ to simultaneously have two properties. First, the projections $f_{\theta_n}(\tilde{x}_k^o)$, on average, should be approximately equal to $f_{\theta_n}(x_k)$  $\forall f_{\theta_n} \in \Omega_{Tr}$. Mathematically, for a small constant $\delta$, we need the average of $\mathcal{L}(f_{\theta_n}(\tilde{x}_k^o), f_{\theta_n}(x_k)) \le \delta$, over all $f_{\theta_n} \in \Omega_{Tr}$. Second, we need the solution $\tilde{x}_k^o$ to lie close to random normal $\mathcal{N}^{L\times H\times W}(0,1)$, to make it perceptually incomprehensible.

Let's consider Eq. \eqref{proxy-opt-sum} with the utility loss term alone, i.e. $\lambda=0$. Theoretically, the only guaranteed solution of this objective is $\tilde{x}_k^o = x_k$. However, we hypothesize that for moderate values of $|\Omega_{Tr}|$ and when $f_{\theta_n}$ represents over-parameterized neural networks, there are infinite solutions for two main reasons. 
First, due to over-parameterization, each $f_{\theta_n}$ ends up being a surjective mapping that allows multiple and diverse inputs $\{\tilde{x}_k\}$ to map to the same output $f_{\theta_n}(x_k)$. In addition, many of these solutions (generated images) are semantically unrecognizable to humans and lie far from the space of natural images. This has been previously observed in the literature [32, 33], and is attributed to the locally linear nature of discriminative models, which assign high confidence even to the regions that lie far from training examples in the high-dimensional input space. 
Second, to ensure utility preservation, practically, we only require the projection $f_{\theta_n}(\tilde{x}_k)$ to be approximately equal to $f_{\theta_n}(x_k)$ for the same argmax class prediction. Hence, a larger range of allowed outputs $f_{\theta_n}(\tilde{x}_k) = f_{\theta_n}(x_k) \pm \epsilon$, for some constant $\epsilon$, can be produced by an even larger and diverse range of inputs $\{\tilde{x}_k\}$.

Due to the diversity of multiple solutions satisfying the utility loss term in Eq. \eqref{proxy-opt-sum}, we hypothesize that specifying a favorable regularisation term should allow us to choose images in a desirable region of space. For instance, in order to make images perceptually incomprehensible, we can simply choose the regularisation to be $\text{ max} \{\mathcal{D} (\tilde{x}_k, x_N) -\tau, 0\}$ to force solutions to lie inside a ball of radius $\tau$ around $x_N \sim \mathcal{N}^{L\times H\times W}(0,1)$. Ideally, choosing such a solution should not affect the compliance with the utility loss term, however, as $|\Omega_{Tr}|$ rises, the allowed freedom is reduced due to more constraints in the utility term, which may affect our ability to regularise the solutions. This effect can be observed in the image IP exposure results in Fig. \ref{fig:tradeoff_ip} and qualitative results in Fig. \ref{fig:qual_results}. As seen, for all datasets, practically there exist values of $|\Omega_{Tr}|$ to synthesize images that are perceptually incomprehensible while simultaneously attaining \textit{desirable} utility preservation for unseen seller models. Hence, this proves our hypothesis correct.

\subsection{Hardness of attacking IPProtect} \label{sec:hardness}
In this section, we analyze the feasibility of successful reconstruction attacks against our method IPProtect. The adversary has access to the sanitized dataset $\tilde{D}_B$, along with an auxiliary dataset $D_A$. The aim of the adversary is to reconstruct the original dataset $D_B$. Here, we provide the reasoning that the probability of successful reconstructions is infinitesimally small. Notice that the transformation $e: (x_k, y_k) \mapsto (\tilde{x}_k^o, y_k)$ in Eq. \eqref{proxy-opt-sum} can be thought of as a loss-plus-regularizer formulation:
{\small
\begin{align}
    \underset{\tilde{x}_k}{\operatorname{argmin}} \sum_{n=1}^{|\Omega_{Tr}|} \ell_n(\tilde{x}_k,x_k) + \lambda \hspace{0.2mm} R(\tilde{x}_k) \label{loss-plus-reg}
\end{align}}

where the loss term $\ell_n(\tilde{x}_k,x_k)$ aims to match the projections, such that $f_{\theta_n}(\tilde{x}_k)\approx f_{\theta_n}(x_k) \forall f_{\theta_n} \in \Omega_{Tr}$, and the regularizer can be arbitrarily chosen to prefer certain solutions in the space of infinite solutions, provided $|\Omega_{Tr}|$ is not too large. Out of multiple potential solutions, an adversary has access to only a certain approximate (low but non-zero $\ell_n(\tilde{x}_k,x_k)$) solution $\tilde{x}_k=\tilde{x}_k^o$, and the aim is to reconstruct the original $x_k$. In one possible attack scenario, the adversary tries to minimize $\ell_n(\tilde{x}_k,\tilde{x}_k^o)$ with an aim to find the original image $\tilde{x}_k = x_k$ from the space of multiple equally-likely solutions to the original loss $\ell_n(\tilde{x}_k,x_k)$.
Since the adversary does not have any reference to $x_k$, there is no way of regularising Eq. \eqref{loss-plus-reg} to pick a solution equal to or close to the original $x_k$. Hence, the solution $\tilde{x}_k = x_k$ can only be chosen by a random chance whose probability is infinitesimally small.

In another attack scenario, an adversary can try to directly learn the reverse-mapping $\tilde{e}: \tilde{x}_k \mapsto x_k$ by generating examples with the help of the auxiliary dataset $D_A$. See Section \ref{gan_attack} for more details. However, learning generalizable mapping is challenging as the mapping is too complex to be learned with a finite parametric model. 
This is due to the fact that $\tilde{x}_k$ only retains limited information about $x_k$, particularly the approximate output class information $f_\theta(x_k)$. Hence, it is almost impossible to reconstruct the lost semantic specifics of the original $x_k$ using a finite parametric model. Moreover, for each $x_k$, our algorithm uses a one-time random seed to initialize $x_N$, which further complicates the learning of the mapping due to the time complexity to decrypt the exact seed being used. In the experiments section, we empirically demonstrate the hardness by attempting to learn the mapping with a Generative Adversarial Network (GAN).

\section {Experiments}
In this section, first, we describe the experimental setup that includes baseline methods, datasets, implementation details, and evaluation metrics. Then, we present an experiment demonstrating the empirical validation of statistical generalization proposed in Section \ref{sec:method}. Next, we present an experimental comparison of the security preservation and utility preservation of our approach in comparison to baseline methods. Here, we analyze the experimental results and discuss why our method can achieve superior performance. Additionally, we present qualitative results of our approach for different values of the security parameter $|\Omega_{Tr}|$. Lastly, we present some additional experiments to study the robustness of our method against low-data scenarios and reconstruction attacks.

\subsection{Experimental setup}
\subsubsection{Baseline methods}
To the best of our knowledge, the task of protecting intellectual property exposure is a novel task that has not been explored earlier. Hence, we repurpose related  approaches, which, similar to our task, aim to obfuscate/ mask raw datasets for secure sharing while preserving their utility. Particularly, we choose two state-of-the-art techniques from the private data synthesis \cite{chen2020gs,harder2021dp,cao2021don} and instance hiding \cite{huang2020instahide,liu2020datamix} literature as baseline methods for comparison, which are described below in more detail. Note that, we did not choose cryptographic methods as comparison baselines since these methods are severely limited to specific setups and are not practical due to substantial computational overhead.

The first baseline is DP-MERF \cite{harder2021dp}, which is the current state-of-the-art method for differentially-private dataset synthesis. DP-MERF takes a raw dataset as input and outputs a private sanitized dataset. This input-output pair is consistent with our task, which makes it a relevant baseline method. However, the original purpose of DP-MERF is to aid statistical analysis, like training a machine learning model, rather than preventing it. We use the buyer's task dataset $D_B$ to train a GAN with DP-MERF's officially released implementation, and then sample the sanitized dataset $\tilde{D}_B$ of the same size. We vary the standard deviation parameter $\sigma$ to control the differential privacy parameter  $\epsilon$, which in turn controls the desired privacy and utility.

The second baseline is InstaHide \cite{huang2020instahide}, which averages a set of random images with random weights, followed by the application of a random pixel-wise mask to hide visual details in an image. The parameter $k$ determines the number of random images to be mixed for one encrypted image and $N$ determines the number of encrypted images to be generated per raw image. A higher value of $k$ and a lower value of $N$ achieves higher security due to a lower signal-to-noise ratio \cite{carlini2021private}. We use the cross-dataset version of InstaHide, which, along with the private dataset, also uses a public dataset to provide better security against reconstruction attacks. As proposed in the original paper \cite{huang2020instahide}, we use $2$ images from the private dataset and $k-2$ images from the public dataset for weighted averaging. We keep $N$ fixed as 1 since $N>1$ version of InstaHide does not provide adequate security due to various successful reconstruction attacks \cite{carlini2021private,luo2021fusion}, which can recover high-fidelity images. Where ever possible, we use the settings for the best performance, like ensuring the mixup coefficient $\lambda$ for the original image to be maximum and training the seller models with the recommended parameters and training protocols (in-inference) of InstaHide. However, for consistency and to ensure a fair comparison, we use our own model architecture, which is fixed for all the baseline methods. To control the utility-security tradeoff, we vary the values of $k$, which is the main parameter of Instahide.

\subsubsection{Data Sets}
We run our experiments using a supervised image classification task on three benchmark, computer vision, datasets MNIST \cite{lecun1998mnist}, FMNIST \cite{xiao2017fashion}, and CIFAR10 \cite{krizhevsky2009learning}. For each benchmark dataset, we use the test set (10K examples) for the buyer task dataset $D_B$, and the train set (50K-60K examples) for generating the seller datasets $\{D_{S_i}\}_{i=1}^{M}$. Specifically, a given seller dataset $D_{S_i}$ is generated by injecting a specific amount of random noise in the train set. The noise is added by randomly flipping labels of a specific percentage of examples. Finally, these seller datasets are used to train seller models $g_{\theta^i}$, whose performances depend on the level of noise in the corresponding seller dataset $D_{S_i}$. 
Using this, we train 8 seller models corresponding to a range of target performance capabilities on the buyer task. The specific values of seller dataset noise levels and resulting performance accuracies of seller models are reported in Table \ref{tab:dataset_noise}. Note that, the exact values of the percentage of labels flipped are obtained by exponentiating 2 with the noise listed in the corresponding rows of Table \ref{tab:dataset_noise} multiplied by 100.

\subsubsection{Implementation details} \label{sec:implementation_details}
We implement our method (Algorithm \ref{alg:proxy_gen}) and conduct all experiments in PyTorch. For clarity, the algorithm has been presented on a data-item level in consonance with Eq. \eqref{proxy-opt-sum} in Section \ref{sec:method}. However, in practice, we speed up the data synthesis by running the algorithm for batches of data. Specifically, in line 5 of the Algorithm \ref{alg:proxy_gen}, instead of a single data sample $(x_k, y_k) \in D_B$, we run the algorithm for a batch of samples ${(x_k, y_k)}_{i=1}^{B_D}$, where $B_D$ is the data batch size. Correspondingly lines 10, 12, 13, and 14 are also converted to their batch-wise variants in code by simply using the batched variants of tensor computations in PyTorch. The specific values of the parameters of our method are presented in Table \ref{tab:parameters}. The values of parameters $\zeta, \lambda, \tau$ are decided by manual inspection of synthesized images for one value of $|\Omega_{Tr}|$ and using the same value for all other $|\Omega_{Tr}|$ runs. The value of $|\Omega_V|$, steps break, model batch size, and data batch size are decided based on ensuring statistical stability, computational, and memory constraints. 

We use custom model architecture configurations for our experiments, which for MNIST and FMNIST is a simple 3 layer Deep Neural Network (DNN) with relu activations, and for CIFAR-10, it's a network with 4 Convolutional Neural Network (CNN) layers and 2 DNN layers along with batch norm, dropouts, and relu activations. 
We run each method for various values of their corresponding security parameters to show security-vs-utility tradeoff characteristics. For InstaHide and DP-MERF, we always vary $k\in\{4,6,10,15\}$, and $\sigma\in\{5,10,15,30,40\}$, respectively. Regarding our method, we fix the values for the security parameters $\lambda$ and $\tau$ and only use $|\Omega_{Tr}|$ to control the tradeoff. The range of values used for   $|\Omega_{Tr}|$ are listed in Table \ref{tab:parameters} for all datasets.
For each method, we manually selected the values of security parameters to show the corresponding method's best variation on the tradeoff curves. In order to minimize the variance, we train 10 separate models with random initialization, using each seller dataset, and, always report the average as the final performance of a given seller. Lastly, for computing the training and inference IP, we sample $D_X$ using the training set of datasets.

\begin{table}[t!]
\centering
\caption{Noise exponent values (Noise) and corresponding ground truth accuracies (Acc.) of seller models for all datasets. Each seller model is trained with a dataset whose x\% labels are flipped randomly, where x\% is obtained by exponentiating $2$ with the corresponding noise exponent in the table multiplied by 100. All accuracy values are averaged over 10 models per seller trained from scratch with random initialization. As seen, we select these noise exponents to make sure we have a diverse range of sellers for each benchmark dataset.}
\begin{tabular}{cllllll}
\multicolumn{1}{l}{} & \multicolumn{2}{c}{MNIST}                              & \multicolumn{2}{c}{FMNIST}                             & \multicolumn{2}{c}{CIFAR10}                          \\ \cline{2-7} 
\#Seller             & \multicolumn{1}{c}{Noise} & \multicolumn{1}{c}{Acc.}   & \multicolumn{1}{c}{Noise} & \multicolumn{1}{c}{Acc.}   & \multicolumn{1}{c}{Noise} & \multicolumn{1}{c}{Acc.} \\ \hline
1                    & 0.00                      & \multicolumn{1}{l|}{95.90} & 0.00                      & \multicolumn{1}{l|}{85.10} & 0.00                      & 79.60                    \\
2                    & 2.00                      & \multicolumn{1}{l|}{89.90} & 2.00                      & \multicolumn{1}{l|}{79.10} & 0.75                      & 72.50                    \\
3                    & 2.50                      & \multicolumn{1}{l|}{87.20} & 2.50                      & \multicolumn{1}{l|}{76.30} & 1.25                      & 65.20                    \\
4                    & 3.00                      & \multicolumn{1}{l|}{83.60} & 3.00                      & \multicolumn{1}{l|}{72.90} & 1.50                      & 60.50                    \\
5                    & 3.50                      & \multicolumn{1}{l|}{78.90} & 3.50                      & \multicolumn{1}{l|}{69.00} & 1.74                      & 53.40                    \\
6                    & 4.00                      & \multicolumn{1}{l|}{62.00} & 4.00                      & \multicolumn{1}{l|}{55.90} & 1.90                      & 48.30                    \\
7                    & 4.50                      & \multicolumn{1}{l|}{45.10} & 4.50                      & \multicolumn{1}{l|}{40.60} & 2.00                      & 44.60                    \\
8                    & 5.00                      & \multicolumn{1}{l|}{28.40} & 5.00                      & \multicolumn{1}{l|}{28.80} & 2.25                      & 31.90                   
\end{tabular}
\label{tab:dataset_noise}
\end{table}

\begin{table}[b!]
\caption{Parameters for IPProtect data synthesis.}
\centering
\begin{tabular}{lccc}
                              & MNIST                                    & FMNIST                                  & CIFAR10                                 \\ \cline{2-4} 
$|\Omega_V|$    & 10                                       & 10                                      & 50                                      \\
$\zeta$          & 0.5                                      & 0.5                                     & 0.2                                     \\
$\lambda$        & 0.1                                      & 0.1                                     & 0.1                                     \\
$\tau$           & 1.3                                      & 1.5                                     & 1.65                                    \\
Steps break (T)               & 2000                                     & 2000                                    & 2000                                    \\
Model batch size (B)             & $|\Omega_{Tr}|*0.95$     & $|\Omega_{Tr}|*0.95$    & $|\Omega_{Tr}|*0.95$    \\
Data batch size ($B_D$)              & 5000                                     & 5000                                    & 256                                    
\end{tabular}
\label{tab:parameters}
\end{table}

\begin{figure*}[t]
    \vspace{-1mm}
    \begin{center}
       	\includegraphics[width=0.85\linewidth]{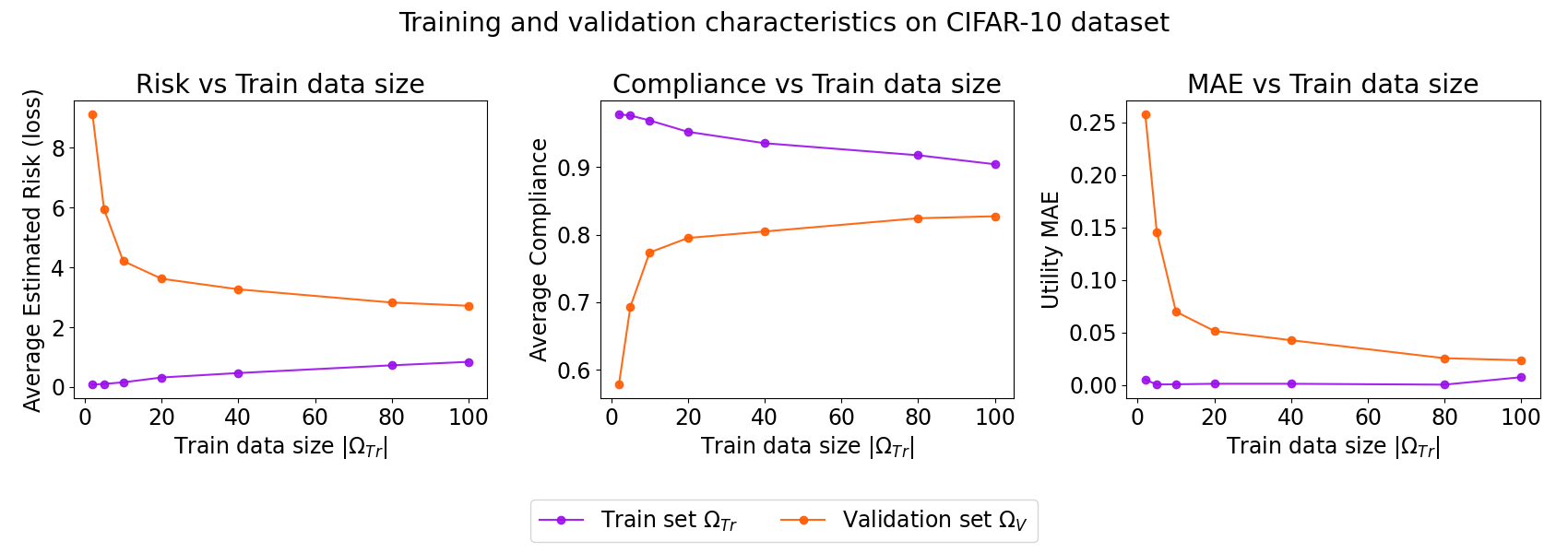}
    \end{center}
    \vspace{-1mm}
\caption{Training and validation characteristics for average estimated risk (left), average compliance (middle), and utility MAE (right), computed on the CIFAR-10 dataset. For the average estimated risk and utility MAE, lower is better, while for the average compliance higher is better. Shows that (1) For higher $|\Omega_{Tr}|$ performance on unseen validation set improves, indicating generalization. (2) A lower data-item-wise risk leads to higher compliance and lower MAE, indicating the effectiveness of our proxy optimization.  (3) For smaller $|\Omega_{Tr}|$, training performance is good but validation is bad, indicating overfitting.}
\label{fig:train_val_charac}
\end{figure*}

\subsubsection{Evaluation Metrics} \label{sec:task_setup_eval_metrics}
For each dataset, the task is to generate a sanitized dataset $\tilde{D}_B$ using the original dataset $D_B$. The performance of all methods is evaluated in terms of two aspects, utility preservation and security preservation, which are governed by an inherent tradeoff. The utility preservation is measured using the Mean Absolute Error (Utility $\text{MAE}$) between the original and the sanitized dataset utilities computed on $M$ seller models. Mathematically, this can be computed as $\frac{1}{M}\sum_{i=0}^M ||U_{g_{\theta^i}} (\tilde{D}_B) - U_{g_{\theta^i}}(D_B)||_1$ where $U_{g_{\theta^i}} (D) = \frac{1}{|D|}\sum_{k=1}^{|D|} (\mathbb{I} (\operatorname{ argmax } g_\theta^i(x_k) == y_k)$ 
and $argmax$ is computed over the C classes. For security preservation, we aim for the lowest possible image and statistical IP exposure of the sanitized dataset $\tilde{D}_B$.  As mentioned in section \ref{sec:prob_deb}, we compute the image IP using Eq. \eqref{def:image_ip} with perceptual loss \cite{johnson2016perceptual}, a popular metric to measure the perceptual distance between a pair of original and its sanitized version. However, associations between original and sanitized images are not explicit in GAN-based approaches. Hence, for DP-MERF, we simply associate a randomly chosen sanitized image with a given original image, albeit, at the cost of giving it some unfair advantage. For computing semantic features, we use features from the $conv5\_1$ layer of the VGG network, which, being a deeper layer is known to correlate better with semantic similarity than a shallower layer \cite{johnson2016perceptual}. Finally, we average perceptual loss across all the images within an entire dataset and report it as the aggregated image IP exposure. Similarly, for statistical IP, we evaluate both training and inference statistical IP, using Eq. \eqref{def:train_stat_ip} and Eq. \eqref{def:infer_stat_ip}, respectively. Also, to summarise the separate aspects of utility preservation and security preservation into a single metric, we compute harmonic mean \cite{okumura2018harmonic} of 0-1 normalized values of utility and intellectual property exposure. Specifically, this is computed as $\frac{x^{-1} + y^{-1}}{2}$, where the utility metric $x \in [0,1]$ and security metric $y \in [0,1]$. For normalization, we simply divide all values of a particular metric by their max value across all baseline methods.

\subsection{Empirical validation of statistical generalisation} \label{sec:stat_generalization}
In this section, we present empirical proof of the concept of novel statistical generalization of synthesized images over a distribution of machine learning models presented in Section \ref{sec:method}. Specifically, first, we simulate i.i.d. sampling of training set $\Omega_{Tr}$, and validation set $\Omega_V$ from an unknown distribution $\Omega_\theta$. Next, we synthesize a dataset $\tilde{D}_O$ from a given original dataset $D_O$ and a training set of models $\Omega_{Tr}$ using the data-item wise ERM optimization Eq. \eqref{proxy-opt-sum}. Finally, we test our hypothesis that the synthesized dataset $\tilde{D}_O$ will generalize to unseen models in $\Omega_V$. Specifically, we show that the estimated risk (and associated metrics) computed using $\Omega_V$ is practically low and results in an approximately similar utility as computed with the original dataset $D_O$. Further, we evaluate the effect of the size of the training dataset $|\Omega_{Tr}|$, which should control the quality of generalization. We explain our experimental setup and present the results to verify our hypothesis below.

Since $\Omega_\theta$ is unknown, similar to a traditional machine learning setup, we simulate i.i.d. samples by simply splitting a large set of sampled models $\Omega_{all}$, into two mutually exclusive sets, training $\Omega_{Tr}$ and validation $\Omega_V$.  In theory, $\Omega_{all}$ may be any complicated distribution, however, in this experiment we use the simple setup where we train machine learning models of a fixed neural network architecture (same as mentioned in Section \ref{sec:implementation_details}) with random initialization and mild ($<10\%$) bootstrapping of the CIFAR-10 train set. Next, as mentioned earlier, we simply sample two mutually exclusive sets of size $|\Omega_{Tr}|$ and $|\Omega_V|$ as the training and validation sets. Finally, using the CIFAR-10 test set as the original dataset $D_O$, we synthesize the sanitized dataset $\tilde{D}_O$ using the data-item wise ERM optimization in Eq. \eqref{proxy-opt-sum}. In particular, we use Algorithm \ref{alg:proxy_gen}, except we do not use our validation set $\Omega_V$ as the stopping criteria, and instead only rely on the fixed number of iterations $T$. This makes sure that we never observe the validation set during the synthesis of the sanitized dataset $\tilde{D}_O$. We use various values of $|\Omega_{Tr}|\in[1, 100]$, while we keep the value of $|\Omega_V|$ fixed to be $10$. 
\begin{table*}[t!]
\vspace{-2mm}
\caption{Shown 0-1 normalized harmonic mean (HMean), along with IP exposure (IP Exp.) and utility MAE (MAE), of all methods on all datasets. We show the average and standard deviation aggregated across all parameter values of a particular method. Note that, in this table format, chosen to allow a better comparison, the same MAE values across IPs are shown multiple times.}
\centering
\resizebox{\textwidth}{!}{
\begin{tabular}{cllllllllll}
\multicolumn{1}{l}{}                                                         &           & \multicolumn{3}{c}{MNIST}                                                                           & \multicolumn{3}{c}{FMNIST}                                                                          & \multicolumn{3}{c}{CIFAR-10}                                                      \\ \cline{3-11} 
\multicolumn{1}{l}{}                                                &           & \multicolumn{1}{c}{MAE} & \multicolumn{1}{c}{IP Exp.} & \multicolumn{1}{c|}{HMean }                  & \multicolumn{1}{c}{MAE} & \multicolumn{1}{c}{IP Exp.} & \multicolumn{1}{c|}{HMean}                  & \multicolumn{1}{c}{MAE} & \multicolumn{1}{c}{IP Exp.} & \multicolumn{1}{c}{HMean} \\ \cline{3-11} 
\multirow{3}{*}{\begin{tabular}[c]{@{}c@{}}Image\\      IP\end{tabular}}     & IPProtect & $\mathbf{0.01\pm .01}$  & $0.71\pm .04$               & \multicolumn{1}{l|}{$\mathbf{0.04\pm .02}$} & $\mathbf{0.01\pm .01}$  & $0.73\pm .04$               & \multicolumn{1}{l|}{$\mathbf{0.04\pm .03}$} & $\mathbf{0.05\pm .02}$  & $\mathbf{0.88\pm .09}$      & $\mathbf{0.17\pm .06}$    \\
                                                                             & DP-MERF   & $0.30\pm .11$           & $1.82\pm .40$               & \multicolumn{1}{l|}{$0.55\pm .08$}          & $0.29\pm .10$           & $1.09\pm .18$               & \multicolumn{1}{l|}{$0.63\pm .09$}          & $0.47\pm .01$           & $1.01\pm .06$               & $0.90\pm .03$             \\
                                                                             & InstaHide & $0.61\pm .00$           & $\mathbf{0.69\pm .01}$      & \multicolumn{1}{l|}{$0.44\pm .00$}          & $0.52\pm .00$           & $\mathbf{0.67\pm .00}$      & \multicolumn{1}{l|}{$0.67\pm .00$}          & $0.42\pm .01$           & $1.19\pm .04$               & $0.92\pm .03$             \\ \cline{2-11} 
\multirow{3}{*}{\begin{tabular}[c]{@{}c@{}}Training\\      IP\end{tabular}}  & IPProtect & $\mathbf{0.01\pm .01}$  & $0.40\pm .07$               & \multicolumn{1}{l|}{$\mathbf{0.04\pm .02}$} & $\mathbf{0.01\pm .01}$  & $0.44\pm .04$               & \multicolumn{1}{l|}{$\mathbf{0.04\pm .03}$} & $\mathbf{0.05\pm .02}$  & $\mathbf{0.10\pm .00}$      & $\mathbf{0.14\pm .05}$    \\
                                                                             & DP-MERF   & $0.30\pm .11$           & $0.40\pm .14$               & \multicolumn{1}{l|}{$0.43\pm .03$}          & $0.29\pm .10$           & $0.41\pm .11$               & \multicolumn{1}{l|}{$0.51\pm .02$}          & $0.47\pm .01$           & $0.10\pm .00$               & $0.47\pm .01$             \\
                                                                             & InstaHide & $0.61\pm .00$           & $\mathbf{0.10\pm .01}$      & \multicolumn{1}{l|}{$0.22\pm .01$}          & $0.52\pm .00$           & $\mathbf{0.17\pm .04}$      & \multicolumn{1}{l|}{$0.37\pm .07$}          & $0.42\pm .01$           & $0.11\pm .01$               & $0.47\pm .01$             \\ \cline{2-11} 
\multirow{3}{*}{\begin{tabular}[c]{@{}c@{}}Inference\\      IP\end{tabular}} & IPProtect & $\mathbf{0.01\pm .01}$  & $0.13\pm .00$               & \multicolumn{1}{l|}{$\mathbf{0.04\pm .02}$} & $\mathbf{0.01\pm .01}$  & $0.18\pm .03$               & \multicolumn{1}{l|}{$\mathbf{0.04\pm .03}$} & $\mathbf{0.05\pm .02}$  & $\mathbf{0.10\pm .00}$      & $\mathbf{0.14\pm .05}$    \\
                                                                             & DP-MERF   & $0.30\pm .11$           & $0.53\pm .13$               & \multicolumn{1}{l|}{$0.51\pm .05$}          & $0.29\pm .10$           & $0.46\pm .14$               & \multicolumn{1}{l|}{$0.54\pm .03$}          & $0.47\pm .01$           & $0.10\pm .01$               & $0.46\pm .02$             \\
                                                                             & InstaHide & $0.61\pm .00$           & $\mathbf{0.10\pm .00}$      & \multicolumn{1}{l|}{$0.22\pm .00$}          & $0.52\pm .00$           & $\mathbf{0.10\pm .00}$      & \multicolumn{1}{l|}{$0.24\pm .00$}          & $0.42\pm .01$           & $0.10\pm .00$               & $0.46\pm .01$            
\end{tabular}}
\label{tab:hmean}
\end{table*}

In order to demonstrate the generalization characteristics, we show results on three metrics for both the training and validation set. The first computes the loss $\frac{1}{|\tilde{D}_O|} \sum_{k=1}^{|\tilde{D}_O|} \hat{R_a}(\tilde{x}_k^o, x_k)$, where $\hat{R_a}(\tilde{x}_k^o, x_k)$ $=\frac{1}{|\Omega_X|}\sum_{n=1}^{|\Omega_X|} \mathcal{L} ( f_{\theta_n} (\tilde{x}_k^o), f_{\theta_n} (x_k))$ computes the empirical risk between a data point $(\tilde{x}_k^o, y_k) \in \tilde{D}_O$ and $(x_k, y_k) \in D_O$ over a set of models $\Omega_X$. The second computes average compliance, which quantifies the percentage of agreeing predictions between the original dataset $D_O$ and the synthesized dataset $\tilde{D}_O$, averaged over all models in a particular model set. Mathematically, we define it as $\frac{1}{|\Omega_X|}\sum_{n=1}^{|\Omega_X|} \hat{C_a}(f_{\theta_n})$, where $\hat{C_a}(f_{\theta_n}) = \frac{1}{|\tilde{D}_O|} \sum_{k=1}^{|\tilde{D}_O|} \mathbb{I} $ $(\text{argmax} f_{\theta_n} (\tilde{x}_k^o) == \text{argmax} f_{\theta_n} (x_k))$ and $\mathbb{I}$ is an indicator function. Lastly, we compute the average Utility MAE, which estimates the absolute difference in utilities of the two datasets, averaged over all models of a particular set. Mathematically, we define it as $\frac{1}{|\Omega_X|}\sum_{n=1}^{|\Omega_X|} ||U_{f_{\theta_n}} $ $(\tilde{D}_O) - U_{f_{\theta_n}}(D_O)||_1$ where $U_{f_{\theta_n}} (D) = \frac{1}{|D|}\sum_{k=1}^{|D|} (\mathbb{I} (\operatorname{ argmax } {f_{\theta_n}}(x_k) $ $ == y_k)$ and $argmax$ is computed over the C classes.

Fig. \ref{fig:train_val_charac} shows the results of the three metrics on the x-axis for different values of $|\Omega_{Tr}|$ on the y-axis. In each plot, we show metrics for both training $\Omega_{Tr}$ and validation set $\Omega_V$. We make several observations. \textit{First}, as the $|\Omega_{Tr}|$ is increased, the performance on the unseen validation set improves, confirming the hypothesis of improving statistical generalization due to better estimation owing to a bigger training data size. Further, as seen in the right-most plot (Utility MAE), for higher values of $|\Omega_{Tr}|$, a small Utility MAE on the validation set indicates that the utility of the synthesized dataset is very close to that of the original dataset in unseen models, empirically validating the utility preservation of the synthesized dataset on unsee models due to statistical generalization. Similar observations can be made in the average compliance (middle) and average estimated risk (left) curve in Fig. \ref{fig:train_val_charac}. \textit{Second}, high compliance of training and validation set demonstrates that minimizing the estimated risk (or loss $\mathcal{L}$) on a data-item level using Eq. \ref{proxy-opt-sum}, indeed results in the same $argmax$ predictions, as hypothesized in Section \ref{sec:method}, in most cases. For instance, a lower loss (estimated risk) for the training set also results in higher compliance of ~90\% in comparison to the compliance of ~83\% for the validation set. \textit{Lastly}, for a smaller $|\Omega_{Tr}|$ the metrics for the training set are much better than those for the unseen validation set, indicative of the overfitting phenomenon owing to the small size of the training dataset $\Omega_{Tr}$. On the other hand, when we increase $|\Omega_{Tr}|$, the training and validation performance approach each other, which is again indicative of statistical generalization.

\begin{figure*}[t!]
    \begin{center}
       	\includegraphics[width=0.85\linewidth]{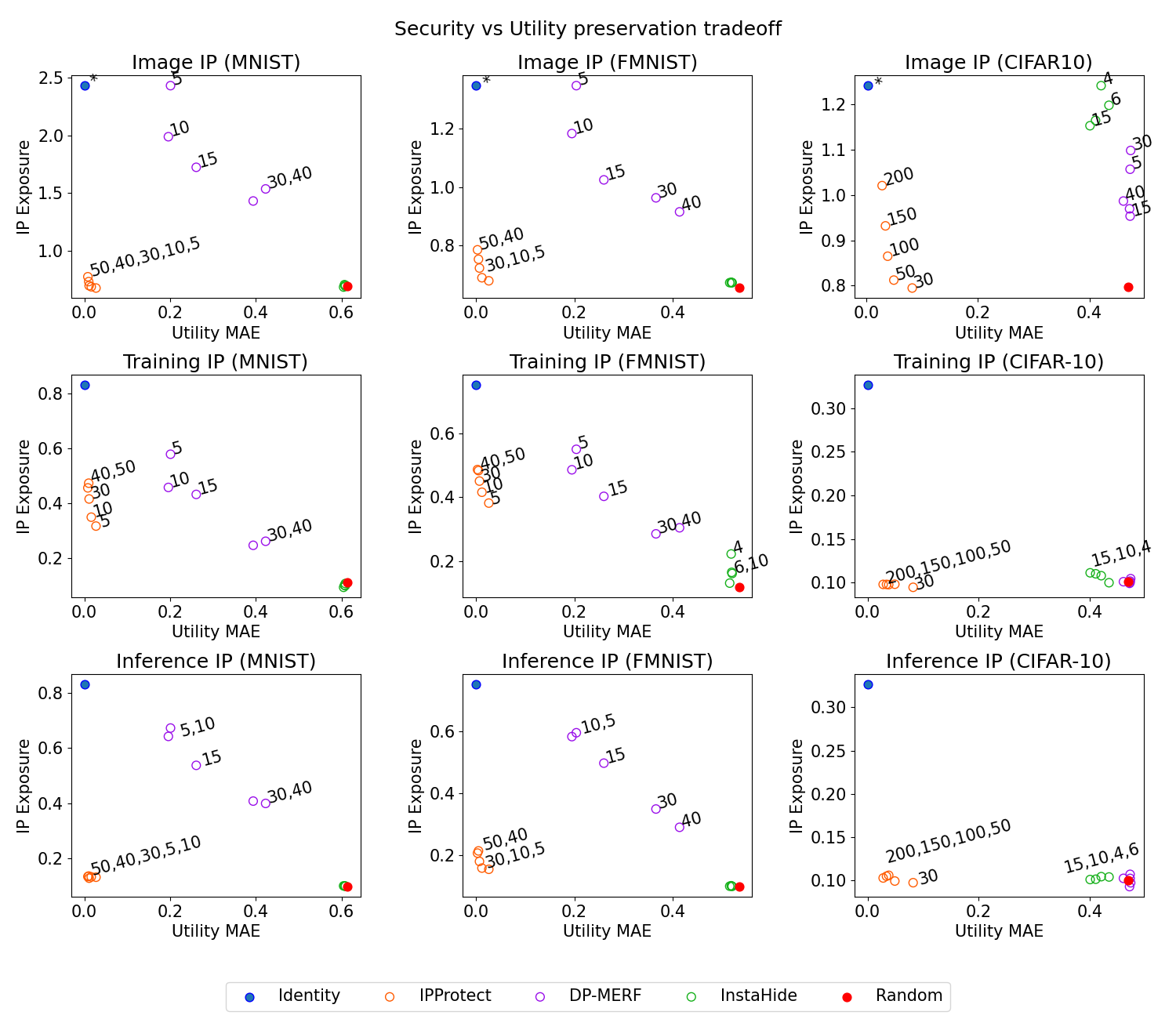}
    \end{center}
\caption{IP exposure vs utility MAE tradeoff curves for MNIST (left), FMNIST (middle), and CIFAR-10 (right) datasets. The top, middle, and the bottom row shows plots for image, training, and inference IP, respectively. For a given method, multiple dots are obtained by varying its corresponding security parameter, resulting in different values of the tradeoff. In some cases, we label the dots with their parameter values to show interesting trends. In the top row, * on the Identity baseline represents that its actual value of IP exposure is $\infty$ and not the one shown. In most cases, IPProtect lies at the lower-left corner, which is the desirable region representing low IP exposure (high security) and low MAE (high utility).}
\label{fig:tradeoff_ip}
\end{figure*}

\subsection{IP exposure experiments}
In this section, we quantitatively compare the performance of all methods in terms of utility preservation and security preservation of the secure data valuation task. First, we present a detailed comparison of various baselines for image IP, followed by training and inference statistical IP. Due to the inherent tradeoff, we present the security-vs-utility preservation tradeoff curves, along with tables containing the average values and 0-1 normalized harmonic mean, defined in Section \ref{sec:task_setup_eval_metrics}. Additionally, we also show qualitative results of our method for different values of the security parameter $|\Omega_{Tr}|$.

\subsubsection{Image IP exposure}
The top row in Fig. \ref{fig:tradeoff_ip} shows the image IP exposure vs utility preservation tradeoff curves for all datasets. In each plot, we show competing methods by varying values of their security parameter, which is $|\Omega_{Tr}|$ for IPProtect (ours), $\sigma$ for DP-MERF, and $k$ for InstaHide. In addition, we also show Identity $\tilde{D_B}=D_B$ and Random $\tilde{D_B}\sim \mathcal{N}^{L \times H \times W}(0,1)$ baselines to give perspective of the two undesirable ends, lacking in either security or utility preservation. Our method lies in the lower-left area, which represents the desired region of high security (low IP exposure) and high utility preservation (low utility MAE). This can be attributed to the joint optimization of utility and security loss in Eq. \eqref{proxy-opt-sum} which aims to preserve projections while forcing images to be similar to  $\mathcal{N}^{L \times H \times W}(0,1)$. As a result, our method achieves utility MAE close to the Identity baseline and IP exposure close to the Random baseline.  Further, we can notice that using a higher $|\Omega_{Tr}|$, akin to a bigger training data size in machine learning, boosts the generalization of statistical estimation, resulting in lower MAE on unseen seller models. Moreover, very good $MAE<0.1$, despite the presence of lower capability sellers, which has a distribution very different from $\Omega_{Tr}$, empirically corroborates the out-of-distribution generalization hypothesis mentioned in Section \ref{sec:method}. 

\subsubsection{Statistical IP exposure} \label{sec:exp_stat_ip}
Fig. \ref{fig:tradeoff_ip} shows the training (middle row) and inference (bottom row) IP exposure vs utility preservation tradeoff curves for all datasets. Similar to earlier, each plot has five baselines and the lower-left area represents the desired region of high utility (low utility MAE) and high security (low IP exposure). As seen, our approach can achieve a better tradeoff in the security-vs-utility curves in all cases, and exclusively lies in the good utility preserving region (MAE $<0.1$). On the security front, our method achieves equal or better IP exposure than all competing methods with reasonable utility preservation ($\text{MAE}<0.25$). This empirically validates our hypothesis that generating images close to $\mathcal{N}^{L \times H \times W}(0,1)$, leads to a dataset with low training and inference IP exposure.

\begin{figure*}[t!]
    \vspace{-4mm}
    \begin{center}
       	\includegraphics[width=0.85\linewidth]{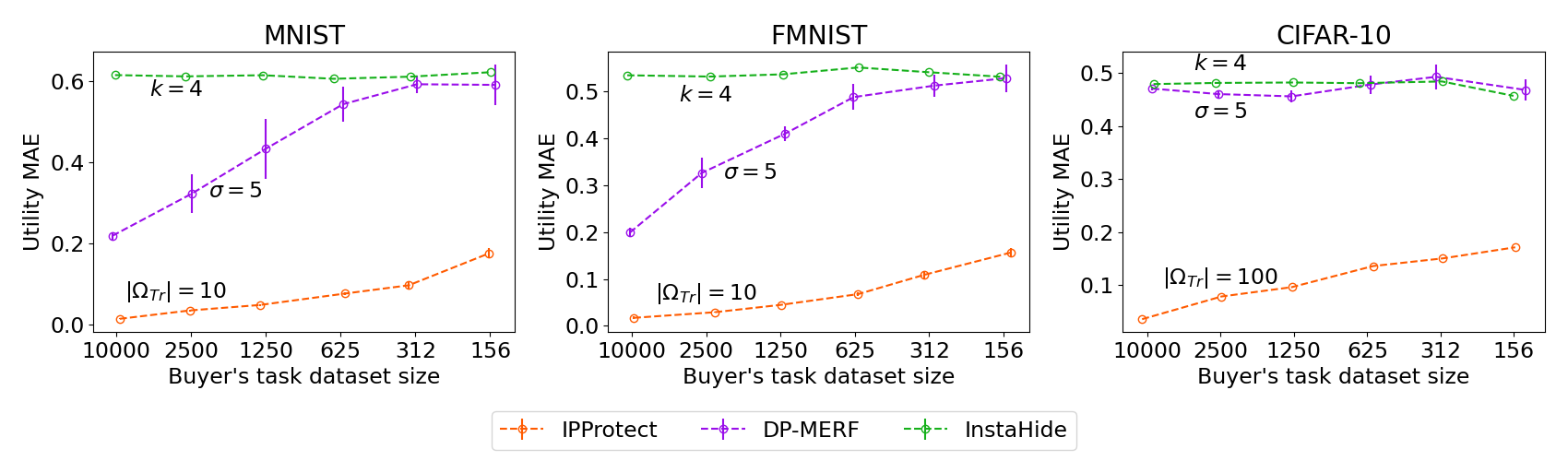}
    \end{center}
    \vspace{-1mm}
\caption{Utility MAE for various values of the buyer's task dataset size for all datasets. Lower values of Utility MAE are better. As seen, MAE gets worse as the dataset size decreases, however, our method, IPProtect continues to outperform all baselines. All results are averaged over 10 random runs with each generating a separate sanitized dataset used for computing these results.}
\label{fig:buyer_data_size}
\end{figure*}

Additionally, Table \ref{tab:hmean} presents the aggregated values, across security parameters, of IP exposure, utility MAE, and the normalized 0-1 harmonic mean, defined in Section \ref{sec:task_setup_eval_metrics}, for all IP exposures on all datasets. In all cases, we consistently achieve the best 0-1 normalized harmonic mean values, which shows a holistic picture of both utility and security fronts. On a side note, notice that InstaHide's utility MAE performance is close to random, as the employed version with $N=1$, which is resistant to reconstruction attacks, performs poorly during inference, despite using its best possible setting.

\begin{figure}[b!]
\vspace{-3mm}
    \begin{center}
       	\includegraphics[width=0.92\linewidth]{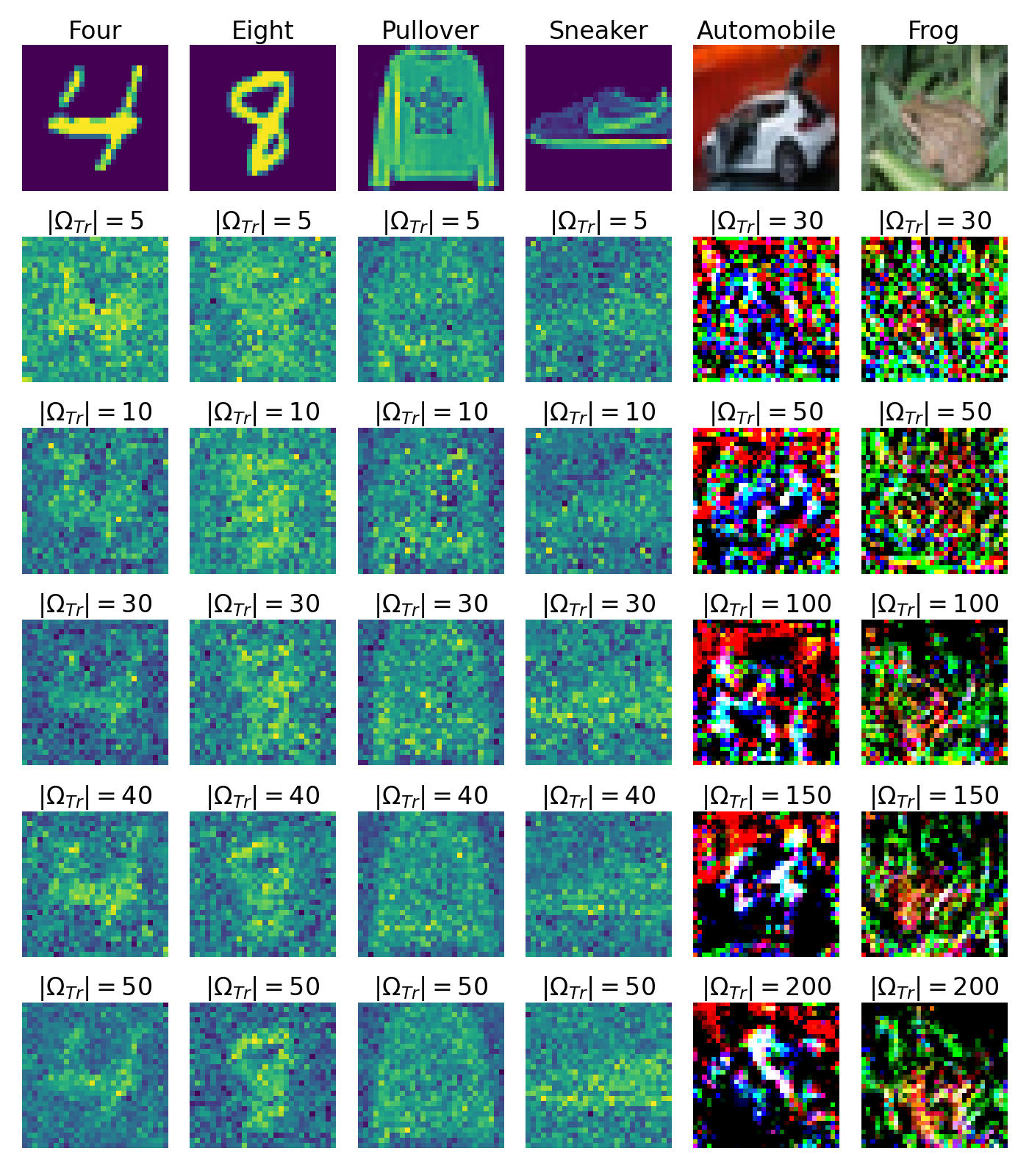}
    \end{center}
    \vspace{-3mm}
\caption{Qualitative results of our method, IPProtect. The first row from the top corresponds to the original images and the following rows show synthesized images for different values of the security parameter $|\Omega_{Tr}|$. From the left, columns 1-2 are for MNIST, columns 3-4 are for FMNIST, and columns 5-6 are for the CIFAR-10 dataset. As shown, the synthesized images are perceptually not as good as their original counterparts, hence, arguably they lost their trade-worthiness, protecting the image IP.}
\label{fig:qual_results}
\end{figure}

\subsubsection{Qualitative results}
Lastly, we show the qualitative results of our method (IPProtect) in Fig. \ref{fig:qual_results} for the corresponding values of the security parameter $|\Omega_{Tr}|$. The synthesized images, especially for our recommended parameters of 10 for MNIST and FMNIST and 100 for CIFAR-10, indeed look like random noise while achieving good utility preservation (Fig \ref{fig:tradeoff_ip}). This effectively protects the image IP as these images are, arguably, no longer tradeable in the market. In some cases, especially for high $|\Omega_{Tr}|$, we do note some artifacts of the real images. This is because, as mentioned in Section \ref{sec:sol_exists}, more number of equations adversely affect the ability to regularise solutions to look similar to $\mathcal{N}^{L \times H \times W}(0,1)$. However, unlike privacy, the presence of some artifacts is, arguably, not a deal-breaker for intellectual property security.

\subsubsection{Ablation}
The two main components of our approach are the two losses, which are utility loss and security loss in Eq. \eqref{proxy-opt-sum}. These losses control the weight of the ‘utility preservation’ and ‘security preservation’ of the synthesized datasets, which being conflicting goals result in a tradeoff. A simple ablation by removing one loss (setting its weight to 0) results in uninteresting trivial solutions, which are Random and Identity baseline in Fig. \ref{fig:tradeoff_ip}. Hence, a better ablation is changing the parameters of the minimization in Eq. \eqref{proxy-opt-sum} to control the tradeoff. Although the tradeoff is controlled by various parameters of the minimization term, which are $|\Omega_{Tr}|$, $\tau$ and $\lambda$, their resulting effect on the utility and security preservation is similar. Hence, we primarily focus on changing $|\Omega_{Tr}|$ to control the tradeoff, and the corresponding effects are shown in Fig. \ref{fig:tradeoff_ip} and qualitative results are shown in Fig. \ref{fig:qual_results}.

\subsection{Robustness experiments} \label{sec:robust_exp}
In this section, we present some more experiments to study the robustness of our method. First, we discuss how the performance of our method is affected by the size of the buyer's task dataset. Then, we demonstrate the robustness of our method by running a reconstruction attack using a Generative Adversarial Network (GAN).

\subsubsection{Effect of buyer data size on utility}
In Fig. \ref{fig:buyer_data_size}, we show how the utility preservation, as measured by MAE, changes with the amount of buyer data size for all datasets. In each plot, we show our method in comparison to DP-MERF and InstaHide baselines. For our method, we choose our recommended security parameter that balances utility preservation and security preservation. On the other hand, for other baselines, we chose the variants optimized for best utility preservation, giving them some unfair advantage. The specific values of these parameters are shown in Fig. \ref{fig:buyer_data_size}. As the buyer data size decreases, the performance of our method degrades, which is expected since with lesser data size, the distribution of $\Omega_{Tr}$ deviates more from the expected seller model distribution $\Omega_\theta$. Similarly, other baselines suffer from the degradation as well owing to the fundamental loss of information due to smaller data size. Overall, as seen, our method ends up outperforming all baselines with a significant margin for various values of data size. Moreover, the lower error margins of our approach demonstrate its consistency for stable analysis of utility preservation. The performance of DP-MERF, unlike MNIST and FMNIST, is significantly poor for the CIFAR-10 dataset since training GANs with differential privacy is known to fail miserably on high dimensional datasets \cite{jordon2018pate, harder2021dp, chen2020gs}.

\subsubsection{Attack on IPProtect} \label{gan_attack}
\begin{figure}[h!]
\vspace{-2mm}
    \begin{center}
       	\includegraphics[width=0.90\linewidth]{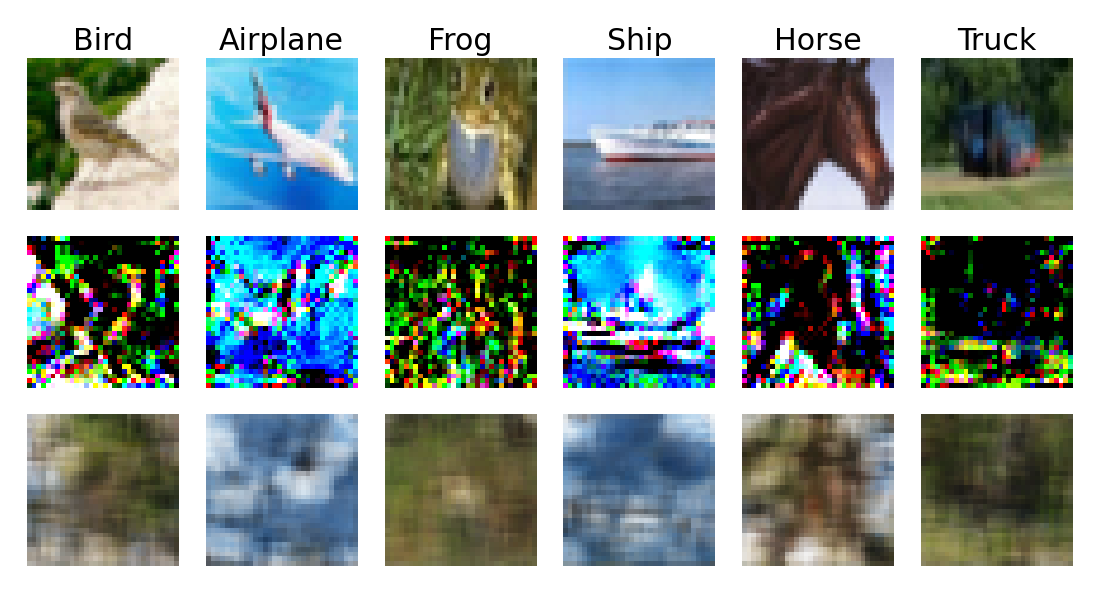}
    \end{center}
    \vspace{-1mm}
\caption{GAN-based reconstruction attack against our method, IPProtect. From the top, rows, (1) show original images, (2) show images generated by IPProtect, and (3) show images reconstructed using a pix2pix GAN. Columns show images sampled from various classes of the CIFAR-10 dataset. As seen, a GAN is not effective in the reconstruction of the original images and sometimes (truck) even predicts the wrong class.}
\label{fig:gan_recons}
\end{figure}

In Section \ref{sec:hardness}, we discussed that given the sanitized dataset $\tilde{D}_B$ and, optionally, an additional auxiliary dataset $D_A$, reconstructing the original dataset $D_B$ is probabilistically hard due to infinite solutions. Here, we empirically demonstrate this claim by trying to learn the mapping $\tilde{e}: \tilde{x}_k \mapsto x_k$ using a GAN. We use the CIFAR-10 train set as the auxiliary dataset sampled from the buyer's task distribution $P_B(X,Y)$. The adversary is assumed to have access to this auxiliary dataset. We, first, create a sanitized version of $D_A$ using our method (Algorithm \ref{alg:proxy_gen}) and denote it by $\Tilde{D}_A$. Then, we train a GAN \footnote{We use this GAN codebase (designed for picture-to-picture conversion, pix2pix, transformation) \text{https://github.com/phillipi/pix2pix}} to generate an image in $D_A$ conditioned on its sanitized version in $\tilde{D}_A$. Finally, we use the sanitized buyer task dataset $\tilde{D}_B$ to generate candidate reconstructions of the original dataset $D_B$. Fig. \ref{fig:gan_recons} shows the qualitative results, which, after extensive hyperparameter optimization, do not look anything like the original images. This suggests that the sanitized images have lost tradeable visual information which is not possible to be reconstructed. On rare occasions (horse, airplane) it appears that the GAN generates an image from the same class, although a completely different image. This happens because the optimization in Eq \eqref{proxy-opt-sum} is designed to lose irrelevant specifics of an image, while only retaining information necessary for predicting the class label.

\section{Conclusions}
In this paper, we tackle the novel task of preemptively protecting the intellectual property of the buyer's task dataset in a utility-based data marketplace. We, first, define novel security risks of image and statistical intellectual property and, then using these definitions, define the objectives of the secure data valuation task. However, solving the secure data valuation task involves intractable optimization. Hence, we propose a proxy optimization that acts as an effective and efficient method to solve the secure data valuation task. We conduct extensive experiments on three computer vision datasets to verify the effectiveness of our approach. Further, we believe our approach has a lot of potential for future research. For instance, a similar idea can be applied for image obfuscation for private edge-cloud inference \cite{liu2020datamix} or potentially for "light-weight" privacy \cite{huang2020instahide, liu2020datamix}.

\bibliographystyle{IEEETran}
\bibliography{main}

\end{document}